\documentclass[runningheads]{llncs}
\usepackage[T1]{fontenc}
\usepackage{graphicx}
\usepackage[colorinlistoftodos]{todonotes}
\usepackage{multirow}
\usepackage{siunitx}
\usepackage{hyperref}
\usepackage{acronym}
\usepackage{cite}
\usepackage{amsmath,amssymb,amsfonts}
\usepackage{algorithm}
\usepackage{algorithmic}
\usepackage{graphicx}
\usepackage{textcomp}
\usepackage{xcolor}
\usepackage{bbm}
\usepackage{bm}
\begin{document}
\title{Explainable Deep Learning Framework for Human Activity Recognition}
%
%
\author{Yiran Huang\inst{1}\orcidID{0000-0003-3805-1375} \and
Yexu Zhou\inst{1}\orcidID{0000-0002-8866-7998}\and
Haibin Zhao \inst{1}\orcidID{0000-0001-7018-1159}\and
Till Riedel \inst{1}\orcidID{0000-0003-4547-1984}\and
Michael Beigl \inst{1}\orcidID{0000-0001-5009-2327}}
\authorrunning{Y. Huang et al.}
\institute{Telecooperation Office, Karlsruhe Institute of Technology, Karlsruhe, Germany\\ \email{\{yhuang,zhou,hzhao,riedel,michael\}@teco.edu} }

\tocauthor{Yiran Huang, Yexu Zhou, Haibin Zhao, Till Riedel, Michael Beigl}

\maketitle              
\begin{abstract}
In the realm of human activity recognition (HAR), the integration of explainable Artificial Intelligence (XAI) emerges as a critical necessity to elucidate the decision-making processes of complex models, fostering transparency and trust. Traditional explanatory methods like Class Activation Mapping (CAM) and attention mechanisms, although effective in highlighting regions vital for decisions in various contexts, prove inadequate for HAR. This inadequacy stems from the inherently abstract nature of HAR data, rendering these explanations obscure. In contrast, state-of-th-art post-hoc interpretation techniques for time series can explain the model from other perspectives. However, this requires extra effort. It usually takes 10 to 20 seconds to generate an explanation. To overcome these challenges, we proposes a novel, model-agnostic framework that enhances both the interpretability and efficacy of HAR models through the strategic use of competitive data augmentation. This innovative approach does not rely on any particular model architecture, thereby broadening its applicability across various HAR models. By implementing competitive data augmentation, our framework provides intuitive and accessible explanations of model decisions, thereby significantly advancing the interpretability of HAR systems without compromising on performance.
\end{abstract}
\newcommand{\vw}{\ensuremath{\mathrm{\boldsymbol{w}}}}
\newcommand{\vb}{\ensuremath{\mathrm{\boldsymbol{b}}}}
\newcommand{\vW}{\ensuremath{\mathrm{\boldsymbol{W}}}}
\newcommand{\vS}{\ensuremath{\mathrm{\boldsymbol{S}}}}
\newcommand{\vx}{\ensuremath{\mathrm{\boldsymbol{x}}}}
\newcommand{\vX}{\ensuremath{\mathrm{\boldsymbol{X}}}}
\newcommand{\vg}{\ensuremath{\mathrm{\boldsymbol{g}}}}
\newcommand{\vc}{\ensuremath{\mathrm{\boldsymbol{c}}}}
\newcommand{\vo}{\ensuremath{\mathrm{\boldsymbol{o}}}}
\acrodef{har}[HAR]{Human Activity Recognition}
\acrodef{sota}[SOTA]{state-of-the-art}
\acrodef{dl}[DL]{Deep Learning}
\acrodef{ml}[ML]{Machine Learning}
\acrodef{cawr}[CAWR]{Cosine Annealing Warm Restarts}
\acrodef{cnn}[CNN]{Convolution Neural Network}
\acrodef{lstm}[LSTM]{Long Short-Term Memory}
\acrodef{da}[DA]{Data Augmentation}
\acrodef{tta}[TTA]{Test-Time-Augmentation}
\acrodef{bn}[BN]{Natch Normalization}
\acrodef{relu}[ReLU]{Rectified Linear Unit}
\acrodef{mlp}[MLP]{Multilayer Perceptron}
\section{Introduction}
The field of Human Activity Recognition (HAR) has seen significant advancements in recent years, driven by the proliferation of wearable devices and the development of deep learning technique. HAR systems, which recognize complex human behaviors from sensor data, have a wide array of applications, from healthcare monitoring to smart home systems. However, as these systems become more integrated into daily life, the demand for explainable AI (XAI) within the HAR domain intensifies. The necessity for XAI stems from a growing need to make the decision-making processes of AI systems transparent, ensuring their reliability and fostering trust among users.

Current methods for providing explanations in AI, such as Grad-Class Activation Mapping (CAM) and attention mechanisms, face significant challenges in the HAR context. These techniques, while effective in visualizing influential data regions for decision-making in image-based models, struggle with the abstract nature of data in HAR. Such data often lack a visual component, making the explanations generated by these methods less intuitive and harder to comprehend. In addition to this, there are some time series interpretation methods such as MCXAI and SBXAI. although they explain decisions through the structural relationships of different cognitive blocks, they are still limited in their exploration of the data, e.g., the effect of data frequency on decision making. 

In response to these challenges, we introduces a simple, novel, model-agnostic framework designed to enhance both the interpretability and performance of HAR models. By employing competitive data augmentation, our approach not only aids in the generation of more intuitive explanations but also does so without the constraints of model-specific architectures. This method allows for a broader application across various HAR models, addressing the limitations of current explanatory techniques.

Our contributions can be summarized as follows:
\begin{itemize}
    \item we propose a unique framework that not only addresses these gaps by offering a model-agnostic solution but also demonstrates how competitive data augmentation can simultaneously improve model interpretability and performance.
    \item Extensive experiments on five benchmark datasets with three state-of-the-art base models validate the effectiveness of the proposed framework.
    \item The source code of OptiHAR has been released \url{http://www.github.com/...}, enabling seamless integration into any given HAR models for boosting their interpretability and performance.
\end{itemize}


\section{Related Work}
Within the study of analyzing sequential data, methods for explaining models without depending on their internal mechanisms can essentially be grouped into three categories, reflecting the core elements they focus on for explanation: instance-based (or feature-based) explanations, subsequence-based, and time-point-based. These categories come with their own unique strategies and challenges.

Instance-based approaches utilize statistical methods to derive features, basing their explanations on how understandable these extracted features are. 

TS-MULE~\cite{schlegel2021ts} evaluates the relevance of segments within the data by forming localized linear models, with techniques like Symbolic Aggregate approXimation (SAX) used for segmenting the data into what can be described as cognitive blocks. SAX-VSM~\cite{senin2013sax} segments the time-series data and constructs a vocabulary of segments, which then helps in explaining the model's input through these defined 'words'. Techniques like MCXAI~\cite{huang2023mcxai} and SBXAI~\cite{huang2023state} delve into understanding the connection between these segments, with MCXAI analyzing spatial connections and SBXAI looking at the temporal flow through different segments. These methods, while insightful, often overlook the deeper temporal aspects and might combine various factors into their explanations, making them complex to grasp.

SoundLime~\cite{mishra2020reliable} introduces slight changes to the original sound files to create new instances. The significance of each moment is evaluated by observing the changes in the model's output for these modified instances. Tsinsight~\cite{siddiqui2021tsinsight} uses a reconstruction method where an auto-encoder, trained on the dataset, attempts to replicate the input, aiding in the explanation process. Salience cam~\cite{zhou2021salience} creates a visual map highlighting important parts of the input by examining how the model's output changes with respect to the input, aiming to pinpoint critical moments. Nevertheless, this method does not adequately address how certain temporal characteristics, like patterns or trends, influence the model, as it mainly focuses on singular points in time.

\section{Methodology}

\begin{figure*}
    \centering
    \includegraphics[width = \textwidth]{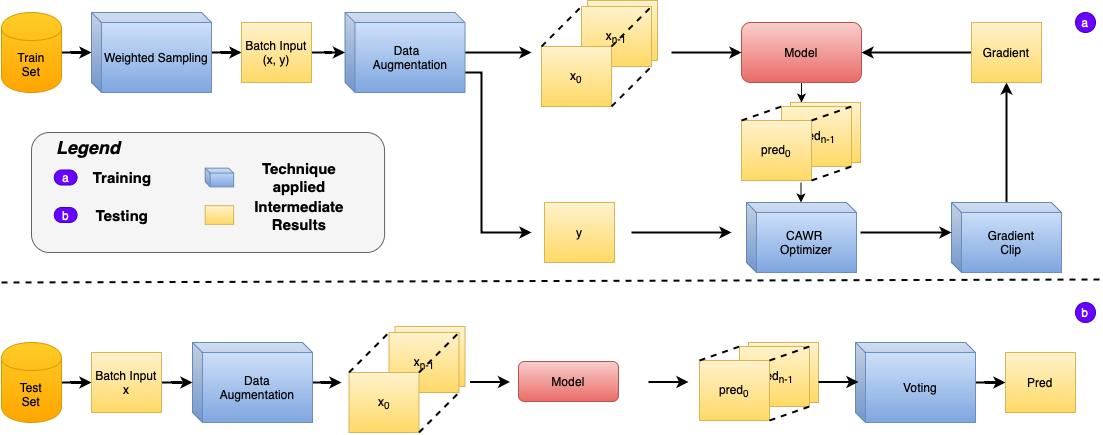}
    \caption{The proposed framework. The blue cubes visualize the techniques that have been implemented such as \ac{cawr}, time series \ac{da}, among others. On the other hand, the orange boxes denote the data flow and the red boxes denote the model.}
    \label{fig:framework}
\end{figure*}

\label{sec: cda}
In this section, we delineate the proposed methodology in detail. Initially, we elucidate the central concept underlying our explanatory framework. Subsequently, we introduce the data augmentation technique employed. Lastly, we detail the comprehensive algorithmic process.

\begin{figure}
    \centering
    \includegraphics[width = .85\textwidth]{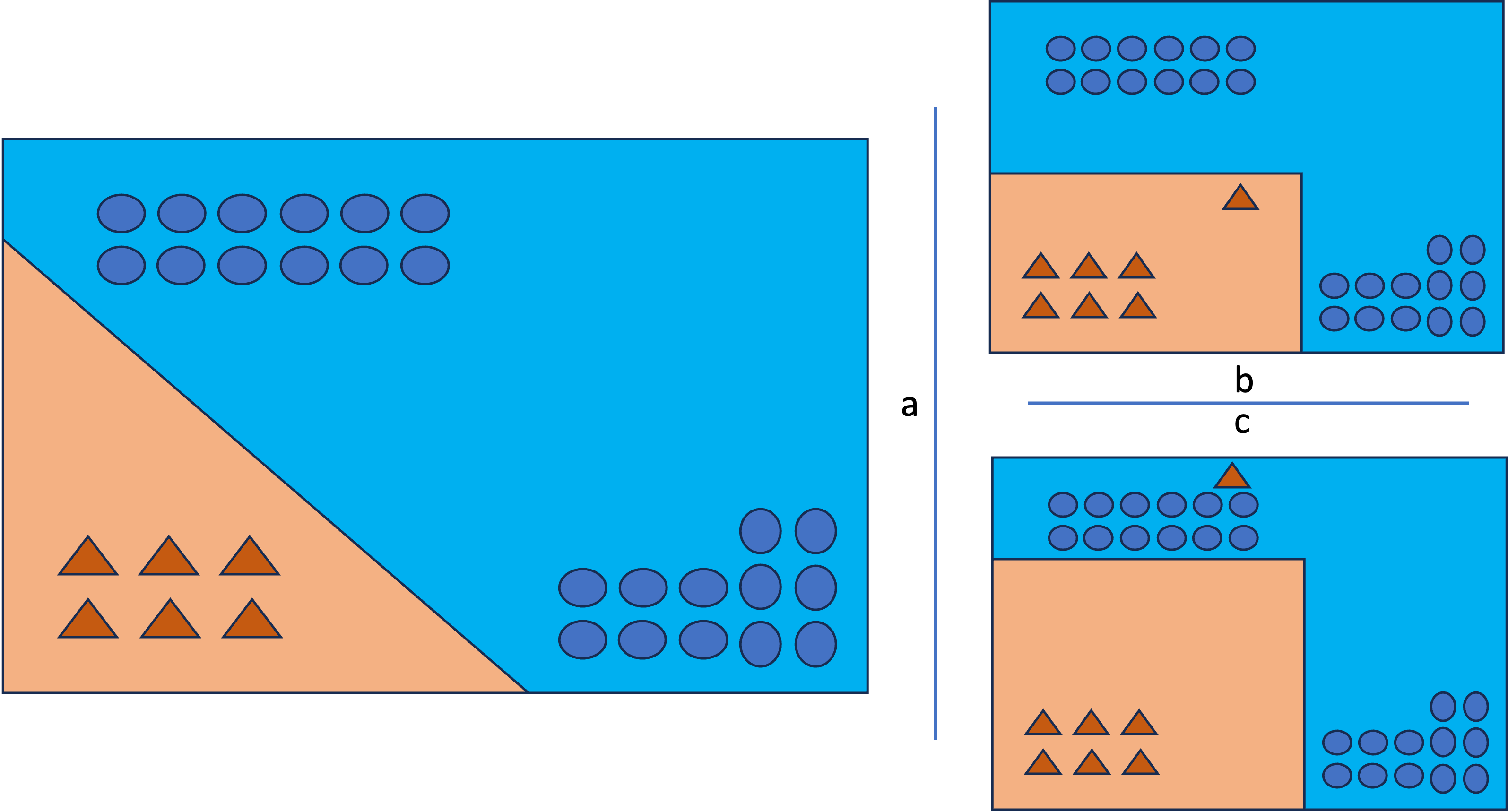}
    \caption{Demonstration of the core idea. Triangles and circles represent different categories, and the lines in the figure depict the boundary lines of the different categories}
    \label{fig:idea}
\end{figure}

\subsection{Competitive Data Augmentation and explanation}
The central element of our proposed framework is the incorporation of data augmentation processes into both the model training and prediction phases. Figure~\ref{fig:idea} illustrates the operational mechanics of this strategy. Data Augmentation (DA) employs predefined transformations to create new data instances that retain the original data's semantic integrity, as outlined by Rumelhart et al. (1986). During the training phase, these transformations are applied to original samples within the dataset to produce variant samples. These variants are then utilized to enhance the model's training, bolstering its robustness and generalizability. As depicted in Figure~\ref{fig:idea}-b, this method refines the model's decision boundaries. In this way, during the prediction phase, samples sharing similar distributions with the augmented variants are classified into the same category as the original samples. However, it is critical to note the scenario depicted in Figure~\ref{fig:idea}-c, where variants distribute near samples from different categories. Although transformations influence the decision boundary, the prediction for a variant may shift during prediction due to a higher sample density from other categories.

The preservation or alteration of model predictions for augmented variants during the prediction phase facilitates the explanation of model decisions. For instance, considering the 'segmentout' transformation depicted in Figure~\ref{fig:da}, if a model's prediction alters post-masking a data segment, that segment is deemed critical for the model's decision. Conversely, if the prediction remains unchanged post-transformation, the segment is considered irrelevant for decision-making. This principle, foundational to counterfactual-based explanation approaches, is innovatively integrated into our data augmentation strategy during prediction. Similarly, changes in model predictions following the addition of high-frequency noise to data imply the significance of high-frequency elements in the original data for predictions.

For the effective implementation of this strategic data augmentation, two conditions must be met: \textit{(i)} The transformations employed during training and prediction must be identical, or the prediction phase transformations should at least be a subset of those used in training. This requirement, which we term 'competitive,' is intuitive. Employing novel transformations during prediction can degrade the model's performance because these new data distributions were not encountered during training. Additionally, it becomes challenging to ascertain whether a change in model prediction is due to the transformed samples' distribution resembling another class or the model's unfamiliarity with this distribution. \textit{(ii)} The quantity of similar distribution samples produced through data augmentation must not exceed the average number of samples per class. This ensures that the augmented samples do not adversely affect the model's assessment of the original samples.

\subsection{Data Augmentation}
\begin{figure}
    \centering
    \includegraphics[width = .85\textwidth]{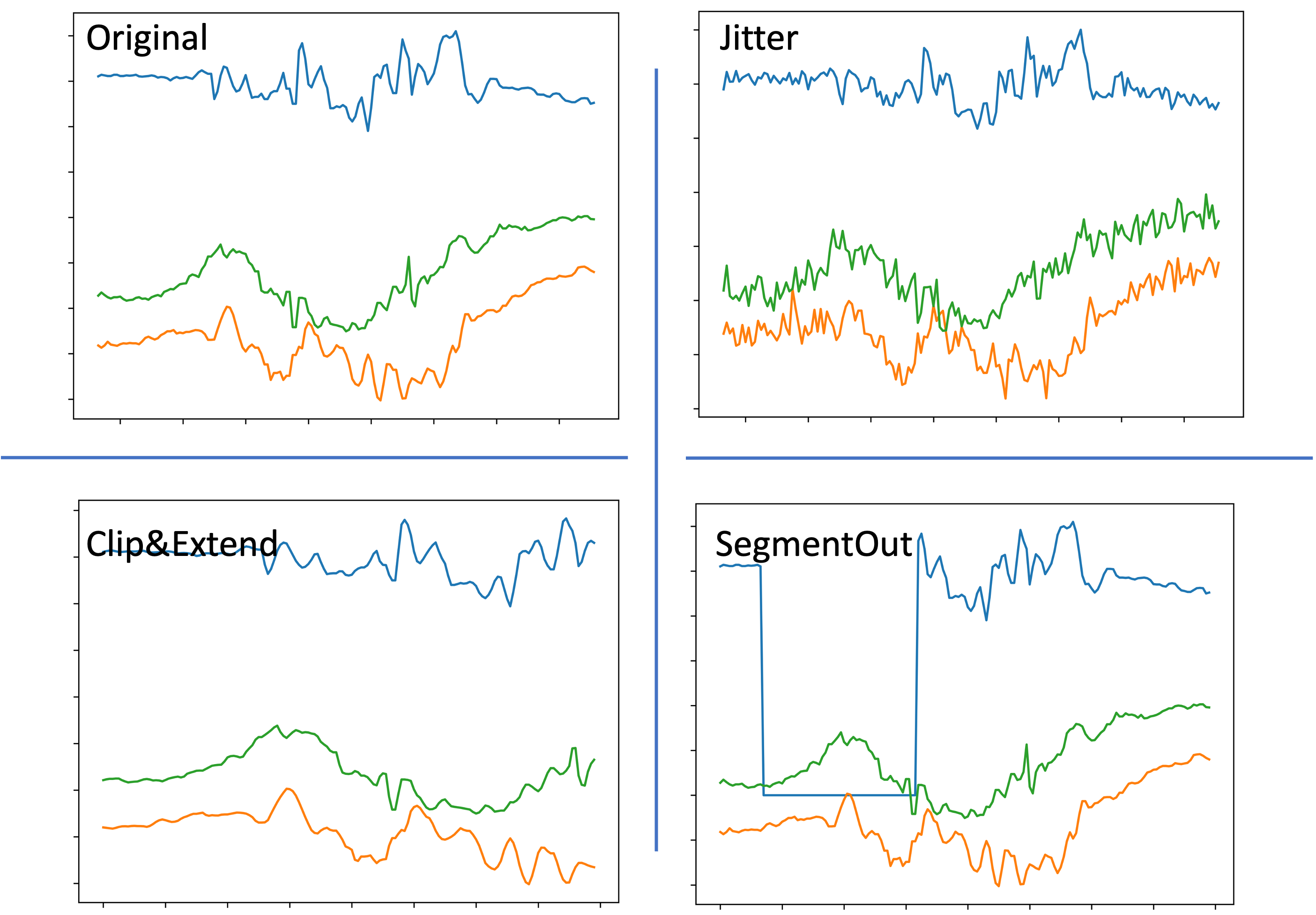}
    \caption{Data Augmentation transformations selected in the framework.}
    \label{fig:da}
\end{figure}

Figure~\ref{fig:da} delineates the data augmentation strategies employed in our study, selected based on two pivotal criteria: realism and comprehensibility. Firstly, we prioritize augmentations that mimic perturbations plausible within real-world scenarios, aiming to ensure that our research aligns closely with practical applications. This not only maintains the augmented data within the model's overall distribution, mitigating additional computational strain,but also potentially elevates the model's performance in realistic settings. Secondly, we emphasize the importance of human interpretability. Given our objective to elucidate model behavior, the value of an explanation diminishes if it is not readily understandable by humans.

\textbf{\textit{Jitter}} generates new data samples by adding random Gaussian noise. This process simulates the noise in the sensor and the real environment.
Recognizing that signals from different sensors possess unique value ranges, we modulate the noise intensity based on the data's variance, formalized as:
\begin{equation*}
    x = x + normal(0, \alpha \cdot \sigma^2),
\end{equation*}
where $\alpha$ adjusts the noise magnitude and $\sigma^2$ is the variance of the sensor's signal. This method allows us to discern the influence of high-frequency and low-frequency signals on model predictions.

Additionally, we simulate data loss scenarios that may occur during data collection or transmission. The \textbf{\textit{Clip}} operation truncates a time segment from the signal, simulating temporal data loss, and compensates for the alteration by interpolating the truncated segment based on the linear model, ensuring consistency with the Human Activity Recognition (HAR) models' fixed input dimensions. Similarly, the \textbf{\textit{SegmentOut}} technique nullifies signals from selected sensors or signal segments, offering insights into the significance of specific numerical segments in driving model decisions. These methodical manipulations facilitate a deeper understanding of data segment relevance in model outcomes.
\subsection{Framework}


%
Figure~\ref{fig:framework} depicts the architecture of the proposed framework, which is bifurcated into training and prediction phases. The framework incorporates a list of transformations for data augmentation. During each iteration of training, a subset of these transformations is randomly chosen and applied to the training dataset, thereby augmenting its size and enhancing the informational depth of the data. This augmented data serves as the foundation for model training. In the prediction phase, a similar approach is employed wherein samples are modified using randomly selected transformations. The model evaluates both the original and the transformed samples. Final predictions are aggregated through a voting mechanism based on the outcomes across these samples. Concurrently, the rationale behind the model's decisions is elucidated by examining the distribution of these votes, providing insight into the model's predictive behavior.
\section{Explanation}
\begin{figure}
    \centering
    \includegraphics[width = .85\textwidth]{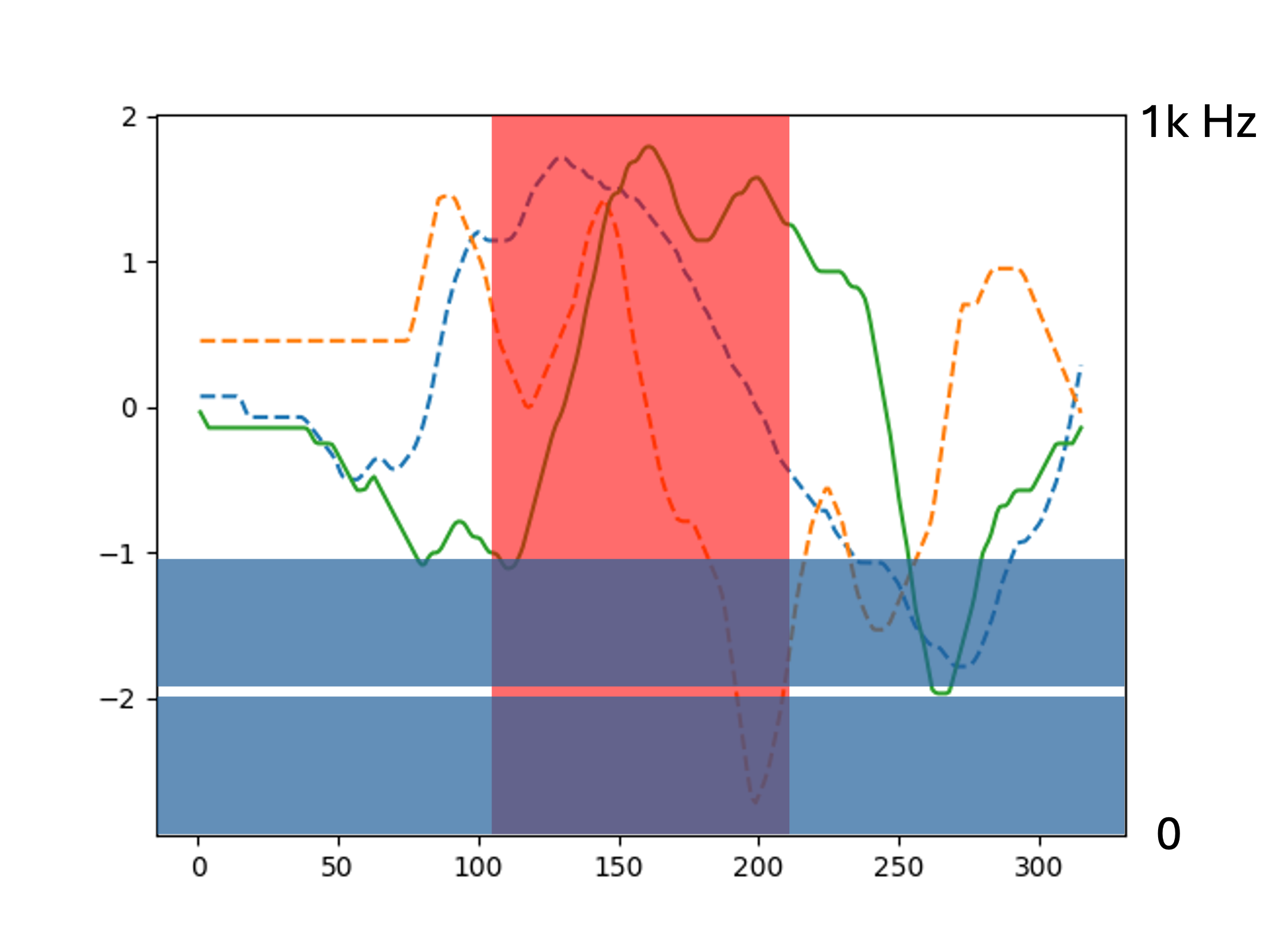}
    \caption{An example of the explanation.}
    \label{fig:explaination}
\end{figure}
In this section, illustrated in Fig. \ref{fig:explaination}, we demonstrate how the proposed approach interprets the model's decisions using an example. The figure includes both solid and dashed lines representing the raw input data, with each line corresponding to the acceleration in the x, y, and z directions. The input data are regularized, and their values correspond to the vertical coordinates on the left side of the figure.

The interpretation provided by the proposed method is determined by the chosen data augmentation techniques. In our experiments, we employed three methods: jitter, clip, and segmentOut. Each method provides a unique perspective on the model's decision-making process.

First, the jitter method adds noise of a specific frequency to the input signal. This allows us to explore the frequency range to which the model is most sensitive. Information within this sensitive frequency domain is considered crucial for the model's decision-making, as the introduction of noise in this domain can alter the model's decisions. In the figure, the blue shaded area represents the sensitive region, with its values corresponding to the vertical coordinates on the right side of the figure.

The clip method involves selecting a portion of the data as input to the model, while the segmentOut method replaces a portion of the data with zeros. Both methods investigate the impact of specific data segments on the model's decisions. In the figure, these effects are illustrated in two ways: Line Type: The dashed lines represent data segments deemed non-essential for decision-making, as the model can make the same decision even without this data. Red Region: The red shaded area indicates the data segments crucial for the model's decisions. Retaining this information alone is sufficient for the model to determine its decisions.

By applying these three data enhancement techniques, we have explored the significance of both regional and frequency domain information in the model's decision-making process. Further insights can be gained by incorporating additional data augmentation methods.

\section{Evaluation}
In the last section, we have demonstrate the explaination of the prediction, in this section, we design two different experiments to answer the following questions about the performance improvement: \textit{(i)}  How well does the proposed method perform compared to the \ac{sota} model-agnostic method? \textit{(ii)} How much does each component of the framework contribute to performance?

In the first experiment, we compare the performance of the given model in the following three scenarios: \textit{(i)} without using the proposed framework (denoted by \textit{Base}); \textit{(ii)} using the entire \ac{sota} model-agnostic method ActivityGAN~\cite{li2020activitygan} (denoted by \textit{activityGAN}); \textit{(iii)} using the proposed framework (denoted by \textit{OptiHAR}).

To detect the contribution of each component, in the second experiment, we compare the model performance in four scenarios: \textit{(i)} without using the proposed framework (denoted by \textit{Base}), \textit{(ii)} using only \ac{da} in the training process (denoted by \textit{DAug}), \textit{(iii)} using only \ac{cawr} (denoted by \textit{CAWR}), and \textit{(v)} using the entire proposed framework (denoted by \textit{Opti}).

\subsection{Benchmark Models}
\begin{figure}
    \centering
    \includegraphics[width = .8\textwidth]{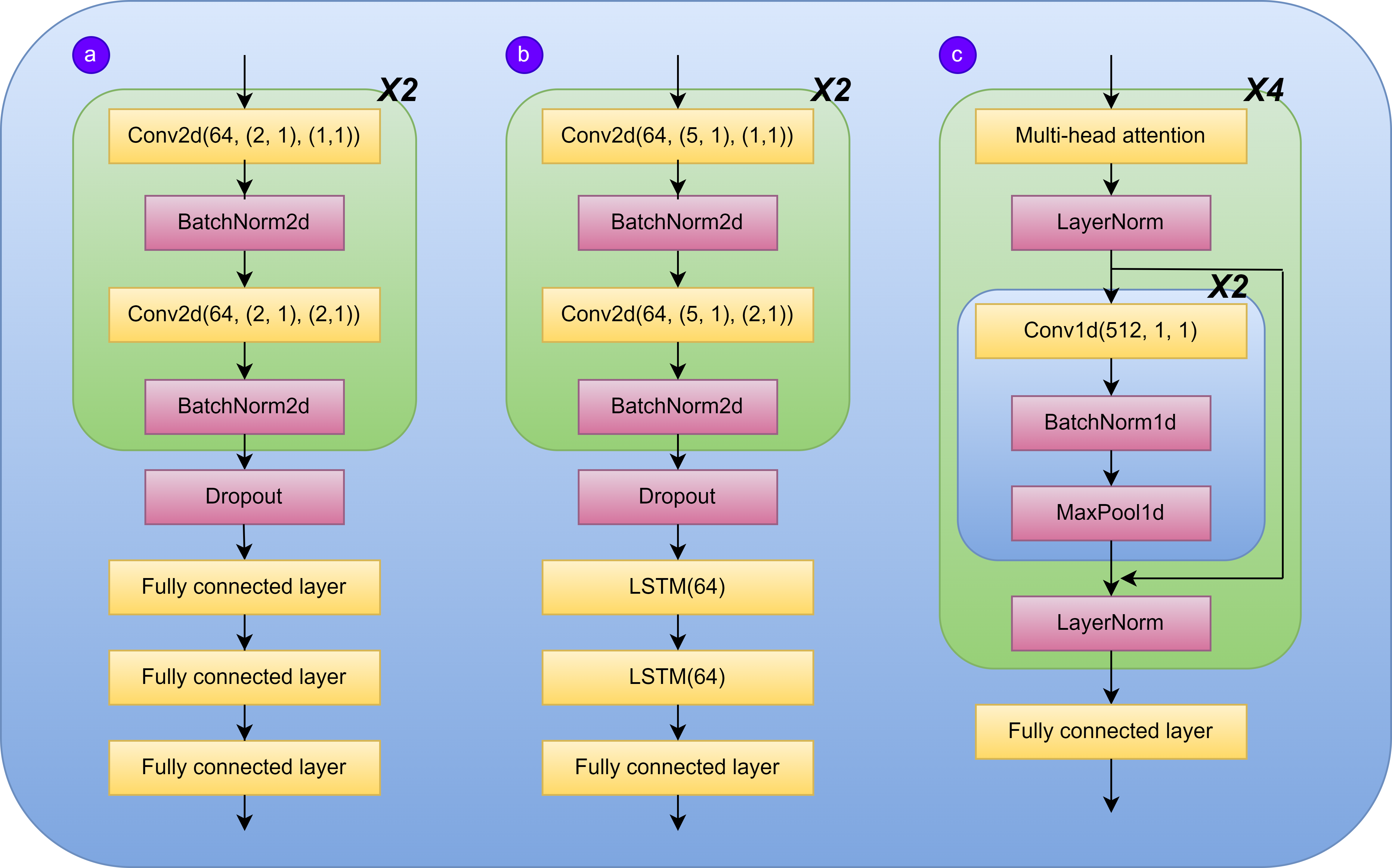}
    \caption{Benchmark model architecture. \textit{a} MCNN. \textit{b} DCL. \textit{c} Transformer. Abbreviations: LSTM, long short-term memory layer; Conv1d, one dimensional convolutional layer; Conv2d, two dimensional convolutional layer; BatchNorm1d, one dimensional batch normalization; BatchNorm2d, two dimensional batch normalization; LayerNorm, layer normalization; MaxPool1d, one dimensional max pooling layer. Parameter: LSTM(hidden dimension number); Conv1d(filter number, kernel size, stride size), Conv2d(filter number, kernel size, stride size).}
    \label{fig:models}
\end{figure}

The \ac{cnn} architecture~\cite{krizhevsky2017imagenet}, \ac{lstm} architecture~\cite{hochreiter1997long}, and Transformer architecture~\cite{vaswani2017attention} are the most widely-used structures in the field of deep learning. Currently, these structures are also being applied in the context of \ac{har}. In order to validate the universality of the proposed framework, we constructed three deep models using these three structures, based on the research by Ito \textit{et al.}\cite{ito2019cnn}, Vaswani \textit{et al.}\cite{vaswani2017attention}, and Zhou \textit{et al.}~\cite{zhou2022tinyhar}. The architecture of each model is presented in \autoref{fig:models}. 

Fig.~\ref{fig:models}(a) shows the \ac{cnn}-based model. It consists of two \ac{cnn} blocks and three \ac{mlp} layers. Each \ac{cnn} block consists of two \ac{cnn} layers and two \ac{bn} layers. Each mapping in the model is followed by a \ac{relu} activation function. Fig.~\ref{fig:models}(b) shows the \ac{lstm}-based model. It consists of two \ac{cnn} blocks, a two-layer \ac{lstm}, and an \ac{mlp} layer. Again, each mapping in the model is followed by a \ac{relu} activation function. Fig.~\ref{fig:models}(c) shows the Transformer-based model, which consists of a \ac{cnn} block and multi-head attention.

\begin{table*}
    \centering
    \caption{Summary of the datasets used in the experiments. The abbreviations acc, gyro, mag denote 3d accelerometers, gyroscopes and magnetometers, respectively.}
    \begin{tabular}{|c|c|c|c|>{\centering\arraybackslash}p{5.3cm}|}
        \hline
         Name & \#Subjects & Sensors type & Freq ($\si{\hertz}$) & Predicted classes\\
         \hline
         \multirow{3}{*}{DSADS} & \multirow{3}{*}{8}  & \multirow{3}{*}{acc, gyro, mag}&\multirow{3}{*}{25}&sitting, standing, walking,\\ &&&& lying, running, exercising, \\ &&&& cycling, rowing, jumping, playing basketball\\
         \hline
         \multirow{3}{*}{HAPT}&\multirow{3}{*}{30}&\multirow{3}{*}{acc, gyro}&\multirow{3}{*}{50}&standing, sitting, lying, walking, walking upstairs, \\&&&&walking downstairs, stand-to-sit, sit-to-stand, sit-to-lie,\\&&&& lit-to-sit, stand-to-lie, lie-to-stand, null\\
         \hline
         \multirow{3}{*}{OPPO}&\multirow{3}{*}{4}&\multirow{3}{*}{acc, gyro, mag}&\multirow{3}{*}{30}&open/close door, fridge, dishwascher,\\&&&& drawer, clean table, \\&&&&drink from cup, toggle switch, null\\
         \hline
         \multirow{3}{*}{PAMAP2}&\multirow{3}{*}{9}&\multirow{3}{*}{acc, gyro}&\multirow{3}{*}{100}&other, lying, sitting, standing, walking, running, cycling, \\&&&& nordic walking, ascending stairs, descending stairs, \\&&&&vacuum cleaning, ironing, rope jumping\\
         \hline
         RW&15&acc&50&jumping, lying, standing, sitting, running, walking, null\\
         \hline

    \end{tabular}
    \vspace{-1em}
    \label{tab:datasets}
\end{table*}

\subsection{Benchmark HAR Datasets}
To test OptiHAR in various scenarios and to keep consistency of the experiments with other works, we employ five widely used benchmark datasets in \ac{har}, namely, DSADS~\cite{data:dsads}, HAPT~\cite{data:hapt}, OPPO~\cite{data:Oppo}, PAMAP2~\cite{data:pamap2}, and RW~\cite{data:rw}. 

\textbf{\textit{DSADS~\cite{data:dsads}.}} DSADS is a dataset that focuses on recognizing daily and sports activities. It includes sensor data from body-worn devices placed at specific locations, such as the wrist or waist capturing movement and orientation during various activities. The sensors are securely fastened to ensure accurate readings while minimizing interference with the participants' movements.

\textbf{\textit{HAPT~\cite{data:hapt}.}} The HAPT dataset utilizes the embedded sensors in smartphones, which are typically carried by participants in their pockets or attached to belts. The smartphones are equipped with accelerometers and gyroscopes to capture the movements and orientations of the users' bodies during various activities. It is designed for addressing issues regarding the occurrence of transitions between activities and unknown activities to the learning algorithm.

\textbf{\textit{OPPO~\cite{data:Oppo}.}} This dataset is aimed at recognizing activities of daily living (ADL) with inertial measurement unit worn at multiple locations on the participant body such as wrists, ankles, and chest using straps or adhesive patches. This setup enables the collection of multi-modal sensor data to capture the body's movements and orientations during activities of daily living.

\textbf{\textit{PAMAP2~\cite{data:pamap2}.}} This dataset includes sensor data from inertial measurement units (IMUs) and heart rate monitors. The IMUs are typically worn on the participant's dominant wrists using straps or bands. The heart rate monitors are worn on the chest, typically utilizing chest straps. This configuration allows for simultaneous measurement of body movement and heart rate during different physical activities.

\textbf{\textit{RW~\cite{data:rw}.}} This dataset focuses on recognizing real-world activities using wearable sensors. It captures sensor data from accelerometers and gyroscopes embedded in smartphones and smartwatches.

These datasets exhibit significant differences in sensor types, mounting locations, sampling rates, and classification activities. To initially prepare the data, we split the data using sliding windows. For the training and validation sets, we employ a $50\%$ overlap between adjacent windows, whereas for the test set, $90\%$ overlap is adopted to  better realistically represent the data slices in the actual execution~\cite{jordao2018human}. A summary of the main information regarding these datasets is presented in~\autoref{tab:datasets}.

\subsection{Experiment Setup}
During the experiment, the parameters $\mathcal{N}_1$ and $\mathcal{N}_2$ of the proposed framework are set to $20$ and $10$, respectively. \Ac{da} parameter  $\alpha$ is set to $0.1$. Clip ratio is set to $0.2$. SensorOut \& SegmentOut ratio is set to $0.1$. The initial repeat period of \Ac{cawr} is set to $50$. $50$ transformation are generted with random parameter to form the transformation set.

\textbf{\textit{Training.}} For all the experiment scenarios, we use an Adam optimizer~\cite{kingma2014adam} with default parameterization and an initial learning rate of $10^{-3}$. Moreover, we employ batch-training with batch size of $256$. As the objective function, cross-entropy loss~\cite{shore1980axiomatic} is utilized for increasing the classification accuracy. We apply early-stopping based on the validation loss and set its patience to be equal to the \ac{cawr} repeat period. The maximal epoch for model training is set to $500$.

\begin{figure}
    \centering
    \includegraphics[width = .7\textwidth]{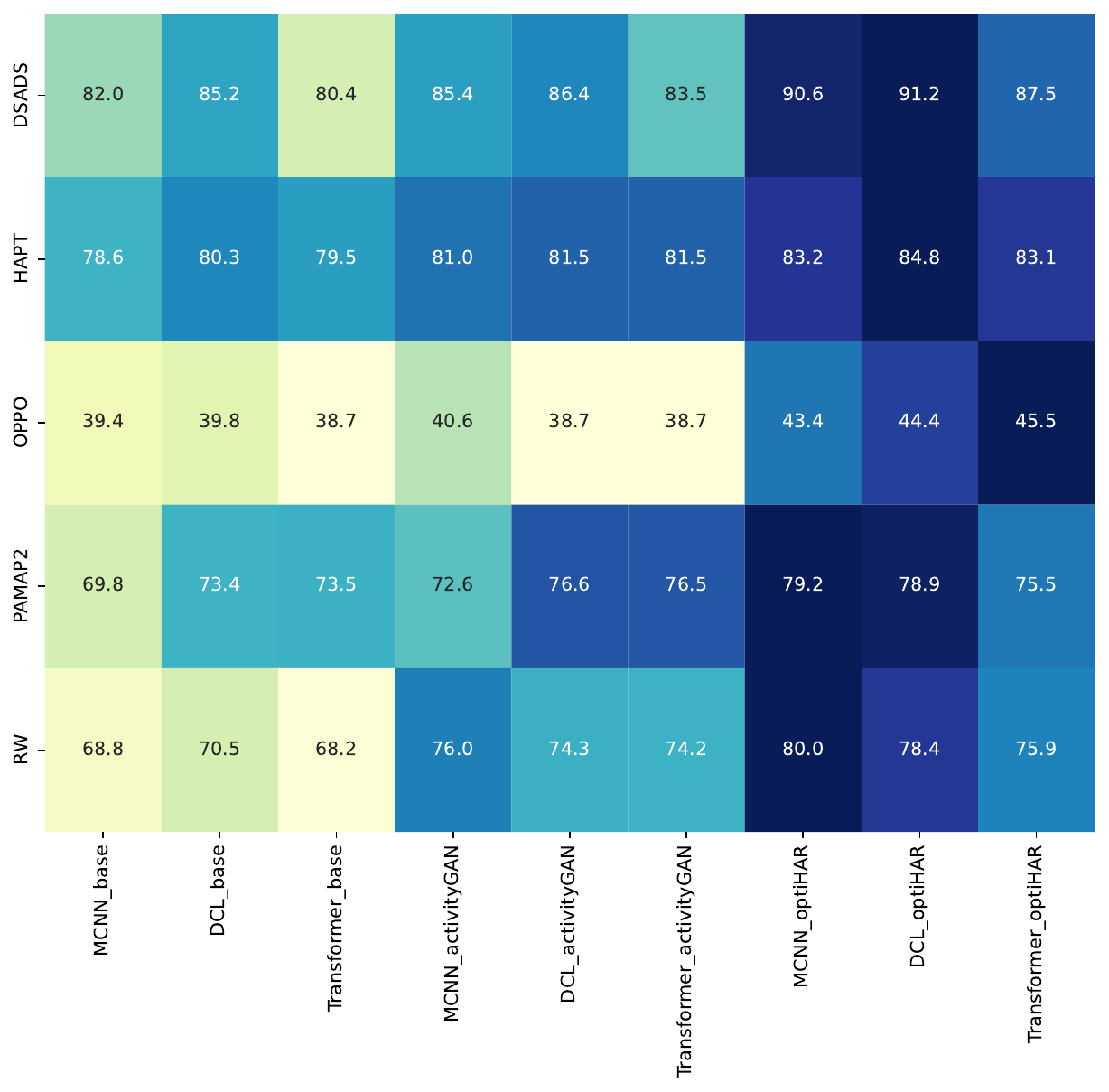}
    \caption{Mean macro F1 scores from LOSO-CV experiments. In each line, the darker the colour, the better the performance of the corresponding method.}
    \label{fig:rsl_exp1}
\end{figure}

\textbf{\textit{Evaluation.}} To examine the generalizability across different subjects, Leave-One-Subject-Out (LOSO) cross-validation (CV) is utilized to evaluate the performance of the trained models. In addition, we employ the macro-averaged F1 score as the evaluation metric~\ref{eq:f1}, which reflects the model's ability to identify each activity without considering the unbalanced distribution of scoring categories. 

\begin{equation}
\begin{aligned}
    F_1 &= (S_p *S_n)/(S_p +S_n)\\
    S_n &= TP/(TP+FN)\\
    S_p &= TP/(TP+FP)
\end{aligned}
\label{eq:f1}
\end{equation}
where TP represents the number of positive instances that were 
classified as positive,
TN represents the number of negative instances that were 
classified as negative,
FP represents the number of negative instances that were 
classified as positive,
FN represents the number of positive instances that were 
classified as negative.

We repeat the experiment five times using different random seeds (ranging from 1 to 5) and report the averaged results w.r.t. the random runs in~\autoref{fig:rsl_exp1} and~\autoref{tab:rsl_exp2}. 

\subsection{Discussion}
The result of the first experiment is summarized in ~\autoref{fig:rsl_exp1}, each row in the figure corresponds to a dataset, and the columns from left to right in the figure correspond to the performance of the specific models (MCNN, DCL, Transformer) under scene basis, activityGAN and OptiHAR. The colours of each row are independent of each other. In each row, darker colors indicate better performance. It is evident that, the proposed framework leads to improved performance across all datasets and models except one (The performance of Transformer model on PAMAP2 dataset). 
The performance of OptiHAR exhibits an average of 4\% improvement on the DSADS, HAPT, PAMAP2 and RW datasets when contrasted with ActivityGAN. Further, on the OPPO dataset, the performance demonstrates a significant improvement of 13\%.

\begin{table}[htbp]
\centering
\caption{Mean macro $F_1$ scores from LOSO-CV experiments. The bold numbers denote the highest F1 score in the corresponding groups.}
\resizebox{\linewidth}{!}{
\small
\begin{tabular}{|c|c|c|c|c|c|c|}
\hline
\multicolumn{2}{|c|}{\textbf{Method}}  & \textbf{DSADS} & \textbf{HAPT} & \textbf{OPPO} & \textbf{PAMAP2} & \textbf{RW}\\
\hline
\multirow{4}{*}{MCNN}&Base&0.837&0.802&0.407&0.701&0.702\\
\cline{2-7}
&DAug&0.834&0.797&0.402&0.700&0.752\\
\cline{2-7}
&CAWR&0.859&0.788&0.392&0.800&0.686\\
\cline{2-7}
&Opti&\textbf{0.906}&\textbf{0.832}&\textbf{0.463}&\textbf{0.829}&\textbf{0.80}\\
\hline
\multirow{4}{*}{DCL}&Base&0.818&0.790&0.381&0.732&0.708\\
\cline{2-7}
&DAug&0.831&0.801&0.378&0.752&0.757\\
\cline{2-7}
&CAWR&0.873&0.803&0.382&\textbf{0.809}&\textbf{0.794}\\
\cline{2-7}
&Opti&\textbf{0.912}&\textbf{0.848}&\textbf{0.474}&0.789&0.784\\
\hline
\multirow{4}{*}{Transformer}&Base&0.804&0.795&0.387&0.735&0.682\\
\cline{2-7}
&DAug&0.795&0.799&0.331&0.730&0.687\\
\cline{2-7}
&CAWR&0.842&0.784&0.382&0.673&0.674\\
\cline{2-7}
&Opti&\textbf{0.875}&\textbf{0.831}&\textbf{0.455}&\textbf{0.755}&\textbf{0.759}\\
\hline
\end{tabular}}
\label{tab:rsl_exp2}
\end{table}

\autoref{tab:rsl_exp2} shows the result of the second experiment. It can be observed that solely employing \ac{da} (which is corresponding to the scenario DAug) did not result in a significant performance gain. \ac{cawr} also results in performance enhancement in approximately half of the instances. 

In addition, compared with other scenarios, the utilization of the proposed framework yields superior outcomes. This could be attributed to the incorporation of other training techniques that enhance the generalizability of the target models.

From the hardware aspect, it is also notable that, this framework does not necessitate stronger GPUs for training, as it doesn't impact the density of the GPU workload, but rather only prolongs its working time due to the slower convergence caused by data augmentation and \ac{cawr}. In execution, e.g., the transformer model on the HAPT dataset takes 0.296 seconds to predict 6629 samples without the framework, while thanks to the parallelization function in PyTorch~\cite{paszke2019pytorch}, the execution time is only increased by $9.8\%$ even using the Opti framework.

\section{Conclusion and future work}

In this study, we present a novel framework designed to enhance both the performance and explainability of models in the field of Human Activity Recognition (HAR). This framework is model-agnostic, allowing it to be applied across various HAR models. At its core, the framework utilizes a competitive data augmentation process that boosts model performance and predictive interpretability through dual-phase data enhancements during training and prediction. Additionally, it integrates several techniques, including Domain Adaptation (DA), bagging, and Class-Aware Weight Regularization (CAWR), in a meticulously crafted manner. This combination improves the overall performance of HAR models without significantly increasing resource requirements. Extensive experiments demonstrate that OptiHAR is versatile and capable of delivering substantial performance improvements across different HAR models.



\bibliography{citation}
\bibliographystyle{splncs04}{}

\end{document}